\documentclass{article}

\PassOptionsToPackage{numbers}{natbib}

\usepackage[final]{neurips_2020}

\usepackage[utf8]{inputenc} 
\usepackage[T1]{fontenc}    
\usepackage{hyperref}       
\usepackage{amssymb}
\usepackage{url}            
\usepackage{booktabs}       
\usepackage{amsfonts}       
\usepackage{nicefrac}       
\usepackage{microtype}      
\usepackage{xcolor}         
\usepackage{times}
\usepackage{amsmath, amssymb}
\usepackage{mathtools}
\usepackage{latexsym}
\usepackage{todonotes}
\usepackage{comment}
\usepackage{tabto}
\usepackage{soul}
\usepackage{multicol,multirow}
\usepackage{graphicx}
\usepackage{verbatim}
\usepackage{authblk}

\def\BibTeX{{\rm B\kern-.05em{\sc i\kern-.025em b}\kern-.08em
    T\kern-.1667em\lower.7ex\hbox{E}\kern-.125emX}}
    
\usepackage{color, colortbl}    
\usepackage{amssymb}
\usepackage{algorithm}
\usepackage[noend]{algpseudocode}
\usepackage{xcolor}

\usepackage{graphicx}
\usepackage{tabularx}
\usepackage{soul}
\usepackage{multirow}
\usepackage{hhline}

\usepackage{epstopdf}
\usepackage[utf8]{inputenc}

\usepackage{hyperref}
\usepackage{xstring}

\usepackage{xcolor}

\author[1,3]{\textbf{Ali Can Kocabiyikoglu}}
\author[1]{\textbf{François~Portet}}
\author[2]{\textbf{Prudence Gibert}}
\author[1]{\textbf{Hervé Blanchon}}
\author[3]{\textbf{Jean-Marc~Babouchkine}}
\author[4]{\textbf{Gaëtan Gavazzi}}

\affil[1]{Univ. Grenoble Alpes, CNRS, Inria, Grenoble INP, LIG, 38000 Grenoble, France  \qquad}
\affil[2]{CHU Grenoble Alpes, Avenue Maquis-du-Grésivaudan, 38700 La Tronche, France  \qquad}
\affil[3]{Calystene SA, 16 Rue Irène Joliot Curie, 38320 Eybens, France  \qquad}
\affil[4]{Clinique de médecine gériatrique, CHU Grenoble Alpes, Équipe Grépi, EA 7408, CS 10217, 38700, La Tronche, France  \qquad}

\title{A Spoken Drug Prescription Dataset in French for Spoken Language Understanding}

\begin{document}

\maketitle

\begin{abstract}
Spoken medical dialogue systems are increasingly attracting interest to enhance access to healthcare services and improve quality and traceability of patient care. In this paper, we focus on medical drug prescriptions acquired on smartphones through spoken dialogue. Such systems would facilitate the traceability of care and would free clinicians' time. However, there is a lack of speech corpora to develop such systems since most of the related corpora are in text form and in English. To facilitate the research and development of spoken medical dialogue systems, we present, to the best of our knowledge, the first spoken medical drug prescriptions corpus, named PxSLU. It contains 4 hours of transcribed and annotated dialogues of drug prescriptions in French acquired through an experiment with 55 participants experts and non-experts in prescriptions. We also present some experiments that demonstrate the interest of this corpus for the evaluation and development of medical dialogue systems.
\end{abstract}

\section{Introduction}

The use of information technology in healthcare has become quite prevalent in the previous years. One of the areas that has largely attracted attention is health dialogue systems used by health professionals, consumers and patients~\cite{bickmore2006health}. Health institutions mostly use Hospital Information Systems (HIS) which have become essential to improve the organization and the quality of care by digitalizing nearly the entire chain of information related to the patient.  One of the major components of an HIS is the Prescription Assistance Software (PAS). 

However, entering information to HIS is time consuming and HIS computers are sometimes far from the point of care. To deal with this situation, we propose to provide a Natural Language interface to the PAS using a smartphone. Such an interface would enable medical practitioners to enter their prescriptions orally at the point of care. Furthermore, this form of interaction would be closer to their usual practice. Such utterance would then be analyzed by Spoken Language Understanding (SLU) to send structured data to a PAS which would validate or not the prescription. Interacting through dialogue would enable the practitioner to enter her prescription quickly while leaving the system some control to make sure no legal information is forgotten (e.g. ``can you specify the duration of the treatment?'').

Dialogue systems, whether they are trained in an end-to-end~\cite{zhao2016towards}~ fashion or in a modular way~\cite{williams2016dialog}, require large amounts of annotated data from both human-human and human-machine interactions, using natural or unnatural or constrained settings~\cite{serban2015survey}.
Even though there is an increasing interest in building systems using publicly available datasets and improving benchmarks for general-domain dialogue systems such as in bAbl tasks~\cite{bordes2016learning}, the distribution of datasets in the biomedical domain is quite limited. For example, the methodical review of ~\cite{wu2020deep} shows that NLP related biomedical research involves a lot of private datasets which are rarely shared or replicated due to patient privacy concerns. This situation is even more difficult for languages other than English which can be subject to different regulations. 

This paper presents PxSLU Corpus\footnote{Our dataset is available at: \url{https://doi.org/10.5281/zenodo.6482586}} , a drug prescription dataset comprising around $4h$ of speech recordings acquired from human-machine interactions using a prototype of a goal-oriented dialogue system used for prescribing medicine. The experiment has been performed in wild conditions with naive participants and medical experts. In total, the dataset includes 1981 recordings of 55 participants (38\% non-experts, 25\% doctors, 36\% medical practitioners), manually transcribed and semantically annotated.  The corpus is made publicly available through a Attribution 4.0 International (CC BY-4.0) license. It is distributed in an aligned format ready for developing and evaluating Spoken Language Understanding (SLU) systems (\textit{conll} format). In this paper, we describe the spoken drug prescription task using smartphones, the data collection protocol and the analysis of the collected data. Furthermore, we present some Natural Language Understanding (NLU) experiments with this acquired data using recent NLP models to show the interest of such dataset for developing NLP technology for health. To the best of our knowledge, the presented dataset is the first corpus of spoken medical prescriptions fully annotated to be distributed to the community. 

\textbf{Outline.}
This paper is organized as following: Section~\ref{sec:contexte} presents the related corpora while Section~\ref{sec:dialogue} introduces the spoken dialogue system which enables medical practitioners to record medical prescriptions through a smartphone. Section~\ref{sec:protocol} explains the data acquisition protocol using our prototype dialogue system. The results of the data collection and their annotation are presented Section~\ref{sec:results} together with experiments comparing recent NLU models trained on external data and evaluated on PxSLU.  Finally, Section~\ref{sec:conclusion} concludes this work and gives some perspectives.

\section{Medical Corpora Related to Prescriptions for NLP}\label{sec:contexte}

\begin{figure*}[tbh]
    \centering
    \includegraphics[width=\textwidth]{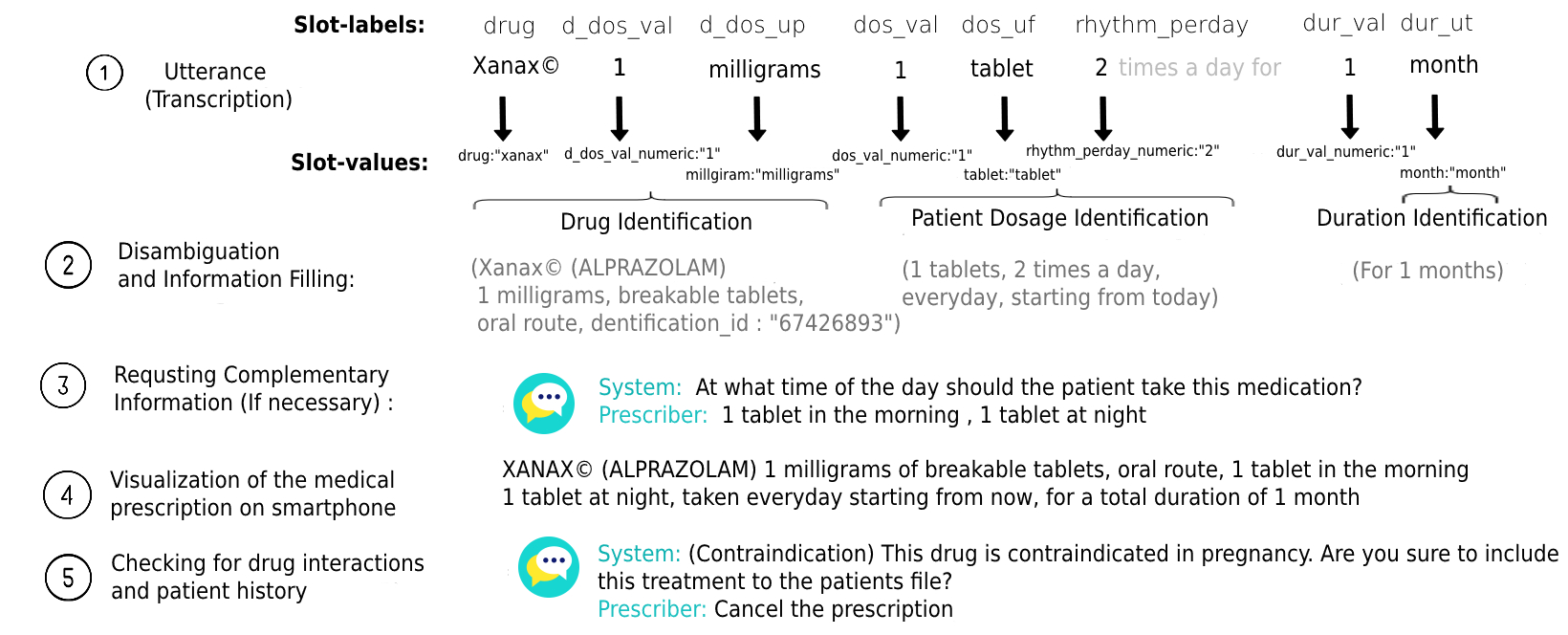}
    \caption{The steps of a dialogue between a prescriber and the dialogue system.}
    \label{fig:etapesGlobales}
\end{figure*}

\subsection{Shared Tasks for Challenges in NLP for Clinical Data}

In the biomedical NLP domain, there are mainly two types of public datasets: first type is the big institutional data warehouses and the second type is the datasets that are collectively built during biomedical challenges and academic datasets for a specific task. 

Most of the biomedical NLP research uses big institutional warehouse datasets such as MIMIC-III~\cite{johnson2016mimic} or AP-HP Health Data Warehouse\footnote{\url{https://eds.aphp.fr/}}. These datasets are distributed as a database with deidentified information and are used for numerous tasks. 

On the other hand, there has been a considerable effort in creating challenges for specific tasks. Such challenges involve and stimulate data collection, annotation, evaluation and tools that are open to the scientific community. The most commonly used challenge datasets are I2B2 (Informatics for Integrating Biology and the Bedside), N2C2 (National NLP Clinical Challenges), and SemEval. Even though most of these datasets are in English, similar challenges exists in other languages than English such as QUAERO corpus~\cite{neveol14quaero} for Named Entity Recognition (NER) in French. The datasets contain texts that are more substantial than simple prescriptions, such as discharge summaries or Electronic Health Records (EHRs). 

\subsection{Drug Prescription Datasets}

Regarding drug prescriptions, we have searched for datasets that could include either whole prescriptions (preferably speech data) and that would contain prescription information available in free text. Although not many, some datasets include drug prescriptions written in natural language. For example, ~\cite{tao2018fable} proposes a semi-supervised prescription extraction system based on information extraction data from medical reports~\cite{uzuner2010community}. There are other challenges that include medical prescriptions mostly in narrative form inside EHRs such as I2B2 medication extraction challenge~\cite{uzuner2010extracting} and Medication and Adverse Drug Events Challenge 2019~\cite{jagannatha2019overview}.  Another source of prescriptions could be MedDialog dataset \cite{chen2020meddiag}. It is composed of 0.26 million medical consultations in English and 1.1 million in Chinese scrapted from online platforms. Each dialogue contains a description of patient’s medical condition, a conversation between a patient and a physician and optionally diagnosis and treatment suggestions. However, the prescription part is not annotated and the dataset is only textual. To the best of our knowledge, there is no dataset of spoken drug prescriptions expressed in a natural way in any language. Therefore we aim to provide to the community a corpus with speech recordings, textual alignments and semantic annotations. 

\section{Understanding Spoken Medical Prescriptions}\label{sec:dialogue}

The Spoken Medical Prescriptions Understanding task follows the work of ~\cite{kocabiyikoglu2019towards}. In this work the Spoken Medical Prescriptions are performed through a cooperative dialogue between a human prescriber and a dialogue system on a smartphone. The starting point of the dialogue is the user who initiates a dialogue session in order to record a drug prescription. The example dialogue shown Figure~\ref{fig:etapesGlobales} illustrates the steps of the dialogue from the initiation to the validation. 

In step \fbox{\textbf{1}}, the system extracts the semantics of the user's utterance  by (\textit{slot-filling}). For each attribute, the system extracts a slot-label, slot-value and a value. For example, the form of the drug destined for the patient is denoted as: \textit{slot-filling} = (dos\_uf,tablet,"tablet"). For more details about the semantic definition of slots, the reader is referred to \cite{kocabiyikoglu2019towards}.

Generally, prescribers' queries are at the same time concise and specific. However, for a system to identify surely a drug, we need to interact with a medical drug database in order to retrieve structured information. For drugs, we distinguish the commercial name from its international non-proprietary name (INN) which is in this case alprazolam. Since a drug can have several entries in a drug database (for example, the same drug can exist as a capsule or a tablet, with different dosages, etc.) it is essential to accumulate all the information about the drug in order to disambiguate and allow its association with a single entry. The step \fbox{\textbf{2}} illustrates this step which allows completing the information in a formal way (Xanax~\textcopyright 1 mg, breakable tablet \dots). If there are several drugs corresponding to the attributes of the current state of the dialog, the system returns a list of drugs proposed to the prescriber. If the system identifies a drug in the users' request, the dialogue continues in order to complete crucial information about the prescription. 

At step \fbox{\textbf{3}}, the system asks the user to complete the dosage (with a valid rhythm and a frequency) and the duration of the prescription. The rhythm of the prescription defines times of the day (morning, noon, etc.) regarding drugs administration whereas the frequency defines an interval in a week (2 times a week).
In our semantic definition, most of the information on the prescriptions is called non-mandatory information.
For example, the fact that a drug should be taken on empty stomach is a good example of a meaningful piece of information for a prescription, but it is optional. In this step, the dialogue management module identifies the mandatory information and associates it with the appropriate dialogue act that will allow it to be collected. Once the dosage, duration and drug information are acquired, the system performs consistency checks on the prescription. In this example, ``twice a day'' is not specific enough, but is an anchor for the dialogue to confirm a valid rhythm and frequency for the prescription.  

The step \fbox{\textbf{4}} concerns the validation of the prescription: the complete prescription will be shown on the screen in a textual from for explicit validation by the practitioner. Once validated, the prescription is sent to the PAS for checking. 

One of the major functionalities of a PAS is its functionality to check for drug interactions and other information related to the drug and according to the patient file. In step \fbox{\textbf{5}}, after adding the prescription, a PAS is able to send back an alert to prescribers specifying the reason for the alert. This step is the last major step in our dialogue system. If the PAS does not find any contraindications, the prescription is added to the patient's record. The prescriber can also validate the prescription to add it to the file after having taken note of any contraindications by validating the return of PAS. In the data collection protocol, the interaction is limited to the validation of the prescription by the user and does not send the data to a PAS. Hence, step \fbox{\textbf{5}} was not used during the data collection experiment. 

\section{Data Collection Protocol}\label{sec:protocol}

The data collection protocol that we design had an objective to collect speech data through human-machine interaction in order to train NLU models and to distribute this data within the community to allow future research and industrial development on this important application.

To allow participants to perform the data collection in their own environment, we deployed a dedicated server in the form of an API that allowed remote participation and data retrieval. This strategy enabled to collect data in a much more ecological way than inviting participants in the lab in order to record interactions in a dedicated room with an experimenter. This strategy also enabled us to perform experiments without breaching the sanitary protocol during the COVID-19 pandemic since participants could stay in their own environment using their own smartphone.  

This required a lot of development and preparation as the participants had to be completely autonomous with their own smartphones. Our goal was to reach about 30 naive users and about 30 experts in drug prescriptions, including some physicians. We have established a simple protocol, inviting the participant to follow the following steps~:

\begin{enumerate}
    \item Registration on the form for requesting to participate in the experiment
    \item Reception of the .apk (installation file of the mobile application on Android) and follow the document explaining the installation and the course of the experiment
    \item Reception of prescription examples (depending on the audience: 20 reading examples for naive users; 10 pictographs and 10 reading examples for medical experts).
    \item Filling the metadata survey (without identifying information), agree to the terms of use, and complete the experiment
\end{enumerate}

Thanks to a fruitful collaboration with the University Hospital of Grenoble, we were able to involve physicians and pharmacists despite the pandemic. Non-expert participants were native French who were either members of the lab of within the social network of the co-authors.

\begin{figure}[htb]
    \centering
    \includegraphics[width=0.35\textwidth]{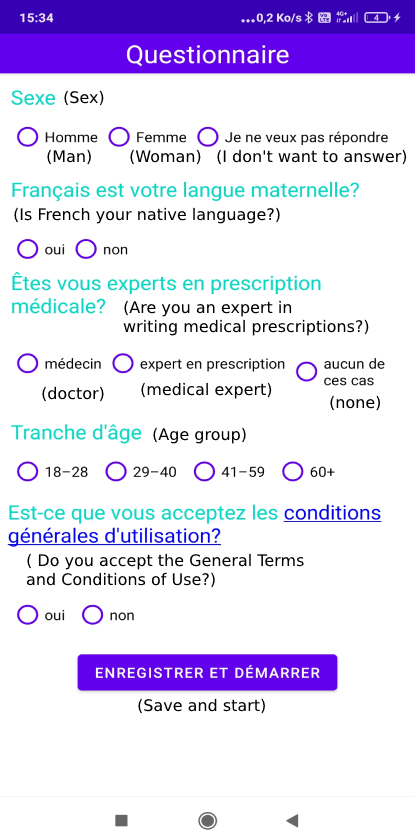}
    \caption{Screenshot of the metadata form of the main interface screen with its English translation (The terms and conditions were accessible by clicking on the link)}
    \label{fig:interfaceMobile}
\end{figure}

Figure~\ref{fig:interfaceMobile} shows the mobile dialogue interface and the demographic information we required. The General Terms and Conditions of use were similar to that of Common Voice\footnote{\url{https://commonvoice.mozilla.org/fr/terms}}, a large-scale effort of the Mozilla Foundation aiming at collect speech from various languages of the planet in order to make it available through a Creative Commons 0 license. The whole experiment has been discussed with the data protection officer of the Grenoble Alpes University and is registered and conforms to the European  General Data Protection Regulation (GDPR).   

\subsection{Data Preparation} 

Since we targeted two types of participants -- medical practitioners (doctors, pharmacists, biomedical engineers, nurses, etc.) and non-experts-- all the participants did not have same expertise on medical prescriptions. Hence we did not provide the same stimuli for these two categories. 

\begin{figure}
    \centering
    \includegraphics[width=0.51\textwidth]{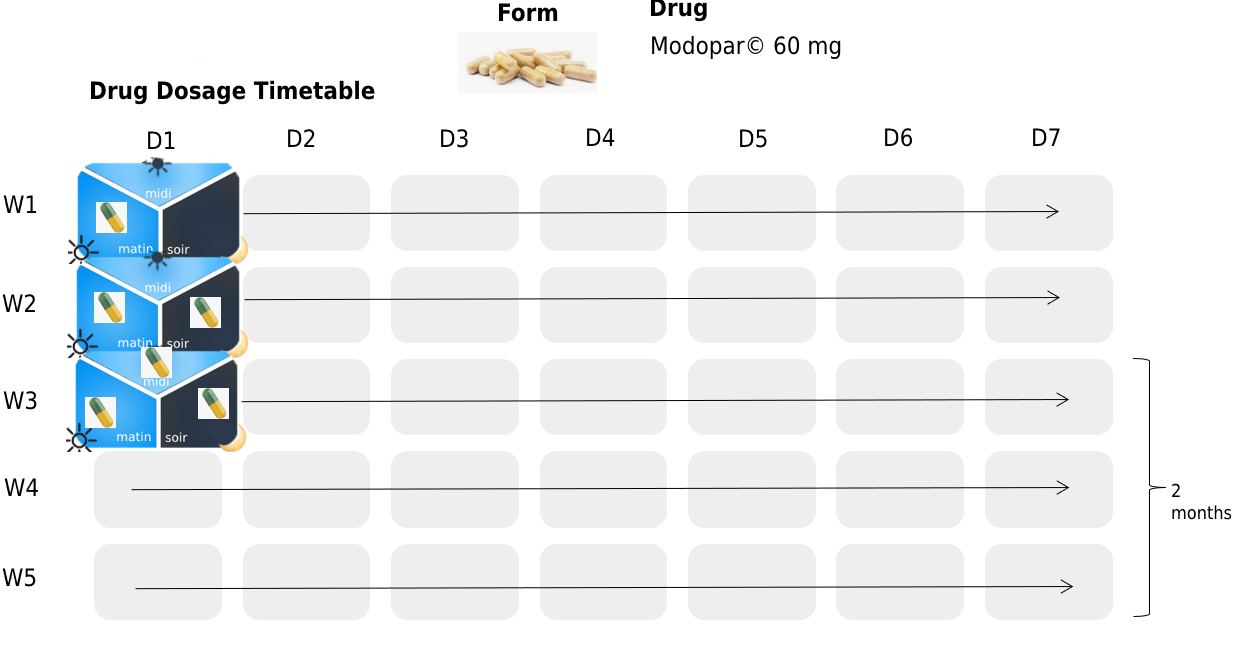}
    \caption{Example of a pictograph representation of a medical prescription}
    \label{fig:pictogramme}
\end{figure}

For non-expert users, we prepared a reading exercise that did not require any domain knowledge. However, for experts we wanted to be as close as possible to their own verbalization. Hence, we defined a method using iconic representations. Indeed, in order to prevent influencing the expert’s utterance, we provided representations of drug prescriptions in the form of diagrams that approximates drug taking timetables. These timetables are designed for patients, generally those who are taking multiple medications, to remind the dosage and times for each medication associated with some conditions and constraints. Figure~\ref{fig:pictogramme} shows an example of this representation. Such graphical representation allows limiting linguistic priors during the experiment.

In Figure~\ref{fig:pictogramme}, the drug (Modopar~\textcopyright) is explicitly given in written text. However, in order not to influence prescribers with the days of the week, the days are represented as (D1,D2,D3,D4,D5,D6 and D7). The dosage form of the drug is represented with an image, which in our example represents capsules but nothing prevents a participant to refer to it as pills or tablets. The names of the drugs, the conditions or the administration details are given in text form at the top of the screen for the prescriber to incorporate into their prescription. The dosage is indicated in the form of a calendar with boxes indicating the start time and their continuity in time. The dosage indicated in the example below denotes the progressive taking of one capsule of the drug in the morning for 1 week, then one capsule in the morning and evening for another week, then one capsule in the morning, noon and evening for 2 months.
Even though this representation allows prescribers to record prescriptions that are more natural, it has the disadvantage of taking more time than a reading exercise. That's why, after 10 pictographs, the experts are asked to perform 10 reading exercises. The non-experts are directly provided with 20 textual stimuli to read. 

To prepare the stimuli, real examples of prescriptions were extracted from books (especially therapeutic books) destined for students in medicine such as \cite{schlienger2013100}, \cite{delcroix2020ordonnances}, \cite{ordonnances180}, \cite{ordonnancesParasitologie}, \cite{delcroix2020ordonnances} and discussed with our experts (two of which are co-authors of this paper). When prescriptions were not complete, we added duration, rhythm and frequency information with plausible values. Given the number of participants targeted, the material generated for the experiment represented approximately 300 examples of pictograph and 1300 textual drug prescriptions. Our preparation included ranking of the prescription according to their complexity (i.e., the longest and the ones with several dosage changes were last).

Based on experience from a previous study~\cite{kocabiyikoglu2020spoken}, the total duration of the experiment was estimated, from setup to data transmission and finalization of the experiment, to 30 minutes. 
\subsection{Recordings Using the Spoken Dialogue System}

The experiment begins with the metadata survey explained in Section~\ref{sec:protocol}. The metadata is saved in a local \textit{sqlite} database in the cache directory of the mobile application. To complete the prescription entry, participants use the ``Push-to-Talk'' button mechanism that triggers the audio recording, and then again to stop the recording. 

\begin{figure}[htb]
    \centering
    \includegraphics[width=0.49\textwidth]{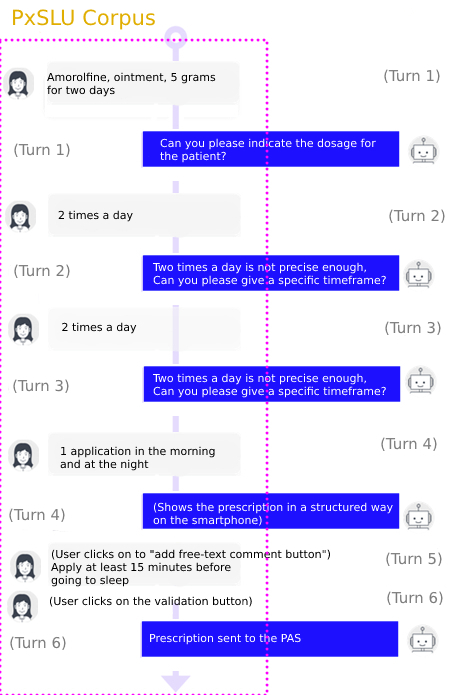}
    \caption{A dialogue example of a dialogue session}
    \label{fig:sampleDialogue}
\end{figure}

Figure~\ref{fig:sampleDialogue} shows an example of a dialogue session recorded with the mobile application. The dialogue session starts with the participant's utterance. The speech local recording is sent to our dedicated server via a secure connection (https). From the server, the recording is then analyzed by Google's automatic speech recognition service. When the server receives the result of the speech recognition (the transcript), it is analyzed by the dialogue system which extracts the intent, semantics and tries to associate the drug-related \textit{slots} with the drug database and determines the continuation of the dialogue which is sent back to the participant's smartphone. The dialogue continues by requesting missing information or responding to the participant's modification requests. A dialogue is completed when the participant validates a complete prescription or cancels it. In the example shown Figure~\ref{fig:sampleDialogue}, the participant gives the rhythm of the prescription as ``2 times a day'' however the system requires a more specific time of the day. 
Then, at dialogue turn \textbf{(4)}, the prescriber gives a more specific time which allows for the system to go one step further and show on the screen the drug prescription in a structured way.

\begin{figure}[htpb]
    \centering
    \includegraphics[width=0.30\textwidth]{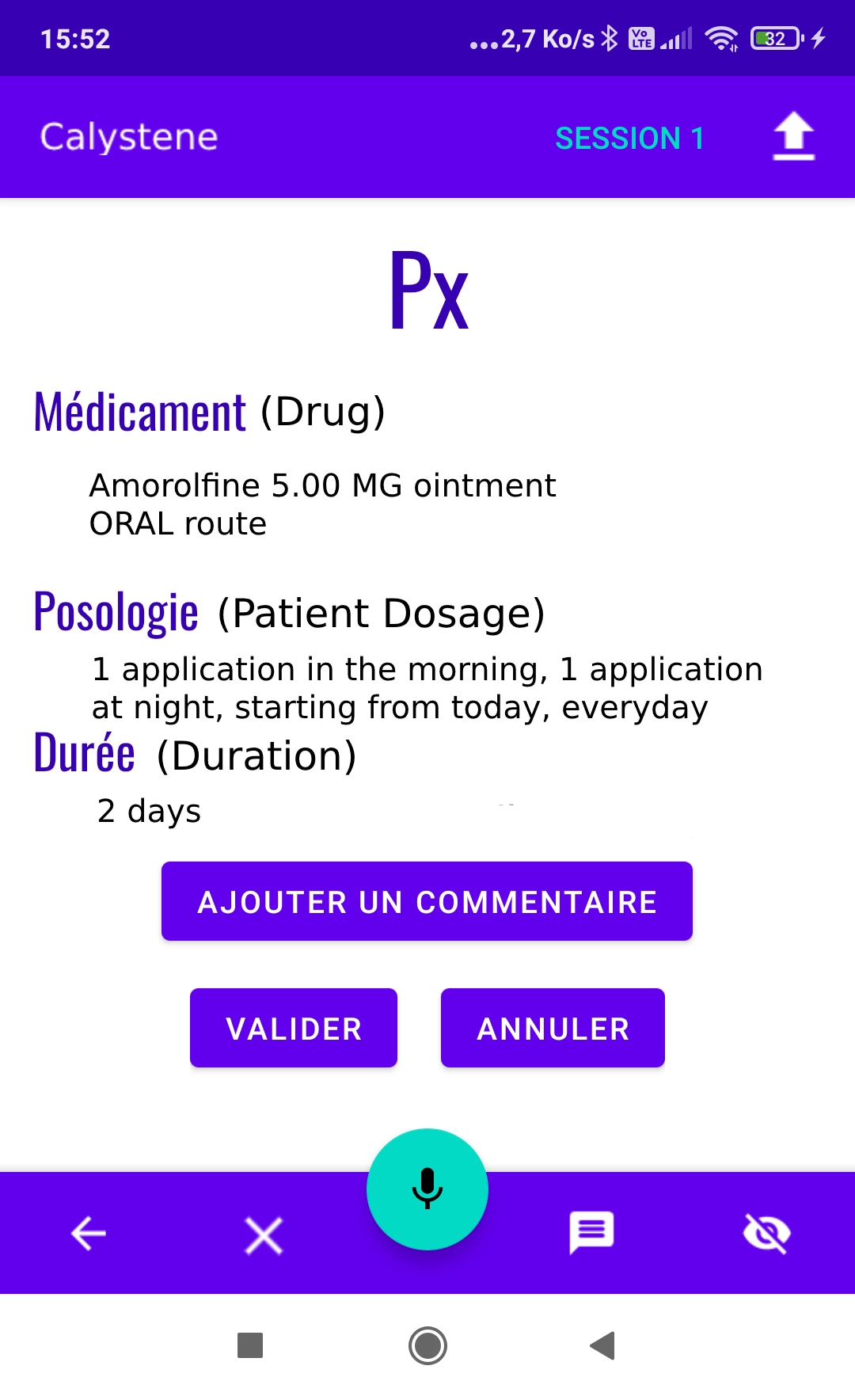}
    \caption{Visualization of a prescription on smartphone}
    \label{fig:affichagePres}
\end{figure}

At the turn \textbf{(5)}, a complete prescription is shown to the participant for validation. Figure \ref{fig:affichagePres} shows the screen capture of the prescription. At this step, if there is a missing or incorrect information, the participant can request to correct the prescription by spoken utterance using the dialogue system. If it's a minor error, or if the participant should add additional non-structured information such as in the example above ``apply at least 15 minutes before going to sleep'', she can click the add a comment button (\textit{ajouter un commentaire}) on \ref{fig:affichagePres} to record the message and include its transcription. Finally, the last step of a dialogue session is when the participant accepts or refuses a prescription by clicking validate (\textit{Valider}) or cancel (\textit{Annuler}). When all of the dialogue sessions are complete, the user can click on the upload button at the top right corner of the screen to  transmit the local database containing statistics and logs and to finish the experiment.

\section{Results}\label{sec:results}

\subsection{Collected Data}

\begin{figure*}[!th]
    \centering
    \includegraphics[width=.8\textwidth]{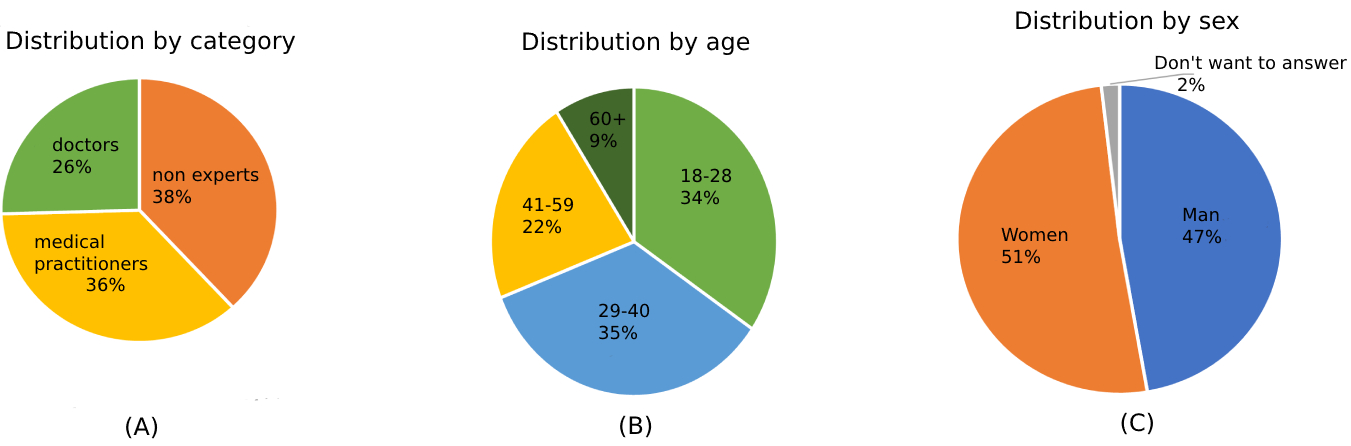}
    \caption{Distribution of some characteristics of the participants}
    \label{fig:repartitionData}
\end{figure*}

The experiment was performed between January 2021 and October 2021 (10 months of experimentation). At the end, 55 databases containing 903 dialogue sessions with 1981 sound recordings were collected.
Table~\ref{tab:statsCollecte} gives a detailed overview of the collected data. The data represents 262 minutes of recordings when all the participants are included. Even though non-experts had initiated more dialogue sessions, total recording time of medical experts and doctors ($\sim$200 minutes) are much more than non-experts ($\sim$62 minutes). This can be explained by the fact that non-experts simply had to read the textual prescriptions and were more likely to exhaust the list than the experts who had a more time consuming interaction with the dialogue system during the phase with pictographs. 

\begin{table}[htb]
    \centering
    {\footnotesize
    \begin{tabular}{|c|c|c|c|} 
\hline
\begin{tabular}[c]{@{}c@{}}\\\textbf{}\end{tabular} & \textbf{Sessions} & \textbf{Recordings} & \textbf{Time (m)}  \\ 
\hline
\textbf{Medical experts}                            & 258               & 434                 & 94.83              \\ 
\hline
\textbf{Doctors}                                    & 230               & 570                 & 105.21             \\ 
\hline
\textbf{Non experts}                                & 415               & 977                & 62.13              \\ 
\hline\hline
\textbf{Total }                                     & 903               & 1981                & 262.27             \\
\hline
\end{tabular}
    }
    \caption{Overview of the collected data}
    \label{tab:statsCollecte}
\end{table}

The data distribution according to several participants' features is presented Figure~\ref{fig:repartitionData}. \textbf{(A)} shows that the collected dialogues are evenly distributed among the participant categories. The pie chart in \textbf{(B)} represents the distribution of the data in relation to age ranges which shows that 3 age ranges are fairly represented while the over 60+ year-old range represents only 10\% of the participants. Finally, \textbf{(C)} shows that the gender representation (M/F) is well balanced.

\subsection{Transcription and Semantic Labeling}

The speech transcription and the semantic annotation of all the dialogues were performed by two native speakers supervised by the co-authors during the summer 2021. 
The two annotators were provided with Transcriber~\cite{barras1998transcriber} and Elan~\cite{hellwig2003eudico} tools and a document describing the transcription convention. Half a day of training was provided by the co-authors. The data consisted in the raw audio recordings and the automatic transcription that was performed by the ASR during the experiment. The task not only consisted in correcting mistakes, but also to obtain transcriptions that are closer to speech utterances. Automatic transcriptions usually remove disfluencies such as repetitions, false starts, etc. but the inclusion of these markers could be advantageous for SLU. The transcription rules made clear all the encountered cases and how to transcribe them. All transcriptions were in lower case and without punctuation. As the dialogue was obtained from human-machine interaction, the dialogue context can influence the transcription process. In order to limit this, and facilitate the transcription process, the corpus was divided into 9 batches of 100 dialogue sessions.Afterwards, these batches were divided equally between the two annotators. 

Semantic annotations were performed using the doccano platform~\cite{doccano} who offers a simple graphical interface enabling to annotate data by clicking and labeling. The labeling process included 5 types of intents and 40 semantic labels that characterize drug prescriptions. Table~\ref{tab:overviewLabels} summarizes the characteristics of the NLU annotations. At the end, 14068 instances of slot-labels and 1981 instances of intents were labeled. The detailed slot-label distribution of the PxSLU corpus could be found in the appendix \ref{annexe:slotLabelDist}. Only 5 slots out of the 40 slot labels had fewer than 12 instances. All other slots had from 5 to 1831 instances.

\begin{table}[htb]
    \centering
    {\footnotesize
    \begin{tabular}{|c|c|c|c|} 
\hline
\textbf{Utterances} & \textbf{Tokens} & \textbf{Slots} &  \textbf{Intents}  \\ 
\hline
1981               & 22440  & 14068                    & 1981                    \\
\hline
\end{tabular}
    }
    \caption{Summary of the semantically annotated PxSLU corpus.}
    \label{tab:overviewLabels}
\end{table}

\subsection{Corpus Analysis}

Transcription errors have a significant impact on SLU systems and the decisions of the dialogue system. For this reason, we first evaluated the word error rate (WER) performance of automatic transcription against reference transcripts. Table~\ref{tab:wer} presents the WER scores for the three categories.

\begin{table}[bht]
    \centering
    \begin{tabular}{|c|c|c|c|} 
\hline
\textbf{}    & \textbf{Experts} & \textbf{Doctors} & \textbf{Non experts}  \\ 
\hline
\textbf{WER} & 21.99\%          & 28.76\%          & 24.42\%               \\
\hline
\end{tabular}
    \caption{WER scores of automatic transcriptions}
    \label{tab:wer}
\end{table}

The WER between the automatic and reference transcripts presented in \ref{tab:wer} is high both for non-experts as for medical experts. The main reason for this difference is related to the labeling convention which included the transcription of onomatopoeias, disfluencies, false starts, \dots 
whereas in the automatic transcriptions, these phenomena were discarded. Also, in the reference transcriptions, numerical expressions regarding drug prescriptions were transcribed all in alphabetic string whereas the ASR used either numerical or alphabetic depending on the context.

\begin{figure}[!bh]
    \centering
    \includegraphics[width=0.45\textwidth]{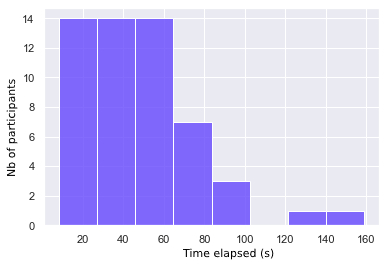}
    \caption{Histogram of average time of session by number of participants}
    \label{fig:histogram}
\end{figure}

In another subsequent analysis, we inspected the elapsed time at a dialogue session level. A dialogue session consists of a user starting a drug prescription until validating or refusing the prescription or restarting the session. Figure~\ref{fig:histogram} shows the histogram of average elapsed time for all of the participants.

Most of the participants completed the task in less than a minute except for a few participants who have a longer dialogue sessions that goes up to 160 seconds, which increases the average elapsed time on the metrics. Our detailed analysis on our three categories of participants show that medical practitioners have spent on average less than 40 seconds in a dialogue session which is a lot less than other categories. This might be explained by the fact that most medical practitioners that participated were pharmacists. As they have more technical knowledge about drugs, the pronunciations were easier for the system to understand. This is confirmed by the lowest WER score presented in Table~\ref{tab:wer}. Furthermore, we found that more specific information about the drugs were given which resulted in reducing the time passed by the slot-filling for missing information. Thus, the validation or the refusal of a prescription was faster.

Additionally, we have noticed that doctors have spent more time on dialogue sessions (avg. 40 seconds - 100 seconds). The average number of dialogue turns per doctors is also higher than other categories. In fact, we have seen that doctors tried to interact more with the dialogue system to correct and add additional information to the visualized prescriptions which also resulted in more validated prescriptions.

\subsection{NLU Model Evaluation}

Our model evaluation builds on previous work \cite{kocabiyikoglu2019towards} where we have presented initial NLU systems trained on artificial and textbook data. 
The training data size was 35676 examples (most of them artificially generated by a context-free grammar). The table \ref{tab:distributionTraining} shows the distribution of these examples by intent.

\begin{table}[!bh]
    \centering
    \begin{tabular}{|l|c|} 
\hline
\multicolumn{1}{|c|}{\textbf{Intent}} & \textbf{Examples}  \\ 
\hline
medical\_prescription                    & 8833               \\ 
\hline
request\_restart                         & 95                 \\ 
\hline
negate                                   & 12608              \\ 
\hline
replace                                  & 12624              \\ 
\hline
none                                     & 1516               \\
\hline
\end{tabular}
    \caption{Distribution of the training corpus}
    \label{tab:distributionTraining}
\end{table}

The NLU models include a classic CRF model, triangular CRF extension (Tri-CRF)~\cite{jeong2008triangular} and Bi-RNN with attention (Att-RNN)~\cite{Liu2016}. Our recent findings on medication information extraction from EHRs in English has shown that transformer-based language models can be extremely competitive~\cite{kocabiyikoglu2021neural}. Even though, there are no available language models trained specifically on biomedical domain, we have decided to include Flaubert, a general-purpose pre-trained transformer language model for French~\cite{le2019flaubert}. 

Table~\ref{tab:resltsNLU} shows the precision, recall and f-measure including micro, macro measures of these NLU systems on PxSLU corpus. The macro average measures are very important since it considers all the equally important \textit{slots} whatever their frequency. Indeed, given the nature of prescriptions, most of the slots are optional and hence occur far less frequently than the mandatory ones.  

The results show that the performance of the model \textit{Flaubert} gives the best results both on micro and macro level performance. All other models have comparable performance. We can see the same behavior for the intent accuracy. However,  and therefore this situation creates an imbalance problem. From the micro average perspective, the Flaubert model is by far the more robust since it performs well even for slots which are rare. 

\begin{table}[bht]
    \centering
    {\scriptsize
    \begin{tabular}{|c|c|c|c|c|c|c|c|} 
\hline
\multirow{2}{*}{\begin{tabular}[c]{@{}c@{}}\\\textbf{ Model}\end{tabular}} & \multirow{2}{*}{\begin{tabular}[c]{@{}c@{}}\textbf{Intent}\\\textbf{(acc)}\\\textbf{ }\end{tabular}} & \multicolumn{3}{c|}{\textbf{Micro Avg}}       & \multicolumn{3}{c|}{\textbf{Macro Avg}}        \\ 
\cline{3-8}
                                                                           &                                                                                                      & \textbf{P}    & \textbf{R}    & \textbf{F1}   & \textbf{P}    & \textbf{R}    & \textbf{F1}    \\ 
\hline
CRF                                                                        & 92\%                                                                                                 & 0.81          & 0.80          & 0.80          & 0.60          & 0.57          & 0.56           \\ 
\hline
Tri-CRF                                                                    & 91\%                                                                                                 & 0.83          & 0.82          & 0.82          & 0.64          & 0.57          & 0.59           \\ 
\hline
Att-RNN                                                                    & 93\%                                                                                                 & 0.83          & 0.87          & 0.85          & 0.55          & 0.55          & 0.53           \\ 
\hline
Flaubert                                                                   & \textbf{94\%}                                                                                        & \textbf{0.89} & \textbf{0.91} & \textbf{0.90} & \textbf{0.69} & \textbf{0.74} & \textbf{0.70}  \\
\hline
\end{tabular}
    }
    \caption{NLU model performance on the PxSLU Corpus}
    \label{tab:resltsNLU}
\end{table}

Apart from providing a real test-bed to evaluate NLU model, we wanted to check if the PxSLU corpus could be used for fine-tuning the Flaubert model. For this purpose, we performed a K-Fold (K=5) cross validation with the pre-trained model ``\textit{flaubert-base-cased}''. In the cross-validation process, the dataset is iteratively split into $k$ roughly equal parts. In each iteration, each of the $k$ part is used as a test set and the rest is used for training. In each run, \textit{fine-tuning} is performed for three epochs. Table~\ref{tab:crossValidation} shows the results of this experiment.

\begin{table}[bht]
    \centering
    {\footnotesize
    \begin{tabular}{|c|c|c|c|c|c|c|} 
\hline
\multirow{2}{*}{\textbf{K \#}} & \multicolumn{3}{c|}{\textbf{Micro Average}}   & \multicolumn{3}{c|}{\textbf{Macro Average}}    \\ 
\cline{2-7}
                               & \textbf{P}    & \textbf{R}    & \textbf{F1}   & \textbf{P}    & \textbf{R}    & \textbf{F1}    \\ 
\hline
\textbf{1}                     & 0.93          & 0.93          & 0.93          & \textbf{0.79} & 0.75          & 0.75           \\ 
\hline
\textbf{2}                     & 0.93          & 0.94          & 0.94          & 0.76          & \textbf{0.80} & \textbf{0.77}  \\ 
\hline
\textbf{3}                     & 0.89          & 0.90          & 0.90          & 0.54          & 0.48          & 0.50           \\ 
\hline
\textbf{4}                     & 0.92          & 0.91          & 0.91          & 0.68          & 0.67          & 0.66           \\ 
\hline
\textbf{5}                     & \textbf{0.95} & \textbf{0.95} & \textbf{0.95} & 0.73          & 0.74          & 0.72           \\ 
\hhline{|=======|}
\textbf{avg}                   & 0.92          & 0.92          & 0.92          & 0.70          & 0.68          & 0.68           \\ 
\hline
\textbf{SD}                    & 0.02          & 0.02          & 0.02          & 0.10          & 0.12          & 0.10           \\
\hline
\end{tabular}
    \caption{K-fold cross validation result of the Flaubert model on PxSLU Corpus (SD=Standard Deviation)}
    }
    \label{tab:crossValidation}
\end{table}

The results shown in \ref{tab:crossValidation} shows that a model trained on PxSLU obtain comparable results with those given in the table ~\ref{tab:resltsNLU}. This shows that PxSLU can be used for fine-tuning and evaluation to lead to similar performance than models training on larger artificial datasets (books as well as artificial data). 
Moreover, the results of the different k-folds vary more when we look at the macro average, which is confirmed by the standard deviation which is also higher. This shows that the choice of data impacts the macro performance and thus the coverage of the \textit{slots}. A finer data partitioning could thus lead to even greater performance. 

\section{Conclusion}\label{sec:conclusion}

We have presented PxSLU corpus which, to the best of our knowledge, is the first drug prescription dataset of speech recordings constructed from human-machine interaction. The dataset includes about $4h$ of speech recordings collected from 55 participants with medical experts. The automatic transcriptions were verified by human effort and aligned with semantic labels to allow training of NLP models. The data acquisition protocol was reviewed by medical experts and permit free distribution without breach of privacy and regulation. The analysis of the corpus and  the evaluation of recent NLU models on PxSLU showed that the dataset is realistic and can be used as a benchmark. Furthermore, it can be efficiently used to fine-tune pre-trained language models.

PxSLU can be used in many other tasks including dialogue \cite{kocabiyikoglu2019towards} and SLU.  We hope that that the community will be able to benefit from PxSLU which will be distributed with a Attribution 4.0 International (CC BY 4.0) license. In a further study, we intend to present check if the dialogue obtained during the experiment can be used to evaluate and train dialogue models.

\section{Acknowledgements}
This work was supported by a CIFRE grant number 2017/1798 from ANRT (National Association for Research and Technology) and was partially supported by MIAI@Grenoble-Alpes (ANR-19-P3IA-0003).

\section*{Appendix: Slot-label Distribution of PxSLU Corpus}\label{annexe:slotLabelDist}

The slot-label distribution of PxSLU corpus is presented in the Table \ref{tab:distributionTab}. As expected from our previous findings~\cite{kocabiyikoglu2019towards}, there is a class imbalance in the semantic information. This is because in drug prescriptions, the mandatory information (drug name, duration, dosage, etc.) are more present than additional information such as (taking on empty stomach).
The slot distribution shows that there are 1285 drug prescriptions (drug$+$inn categories) in the recordings.

\begin{table}[htb]
    \centering
    {\scriptsize
    \begin{tabular}{|c|c|c|c|c|c|} 
\hline
\textbf{drug}                                                              & 1042 & \textbf{inn}                                                              & 431  & \begin{tabular}[c]{@{}c@{}}\textbf{max-}\\\textbf{unit-ut}\end{tabular}    & 22    \\ 
\hline
\begin{tabular}[c]{@{}c@{}}\textbf{rhythm-}\\\textbf{rec-ut}\end{tabular}  & 36   & \textbf{rhythm-tdte}                                                      & 1452 & \begin{tabular}[c]{@{}c@{}}\textbf{min-}\\\textbf{gap-val}\end{tabular}    & 5     \\ 
\hline
\textbf{d-dos-val}                                                         & 954  & \begin{tabular}[c]{@{}c@{}}\textbf{rhythm-}\\\textbf{perday}\end{tabular} & 264  & \begin{tabular}[c]{@{}c@{}}\textbf{min-}\\\textbf{gap-ut}\end{tabular}     & 8     \\ 
\hline
\textbf{d-dos-up}                                                          & 924  & \begin{tabular}[c]{@{}c@{}}\textbf{rhythm-}\\\textbf{hour}\end{tabular}   & 119  & \begin{tabular}[c]{@{}c@{}}\textbf{cma-}\\\textbf{event}\end{tabular}      & 294   \\ 
\hline
\textbf{d-dos-form}                                                        & 303  & \textbf{freq-val}                                                         & 22   & \textbf{fasting}                                                           & 18    \\ 
\hline
\begin{tabular}[c]{@{}c@{}}\textbf{d-dos-}\\\textbf{form-ext}\end{tabular} & 68   & \textbf{freq-startday}                                                    & 7    & \begin{tabular}[c]{@{}c@{}}\textbf{rhythm-}\\\textbf{rec-val}\end{tabular} & 31    \\ 
\hline
\textbf{A}                                                                 & 57   & \textbf{freq-ut}                                                          & 125  & \textbf{re-val}                                                            & 36    \\ 
\hline
\textbf{roa}                                                               & 55   & \textbf{freq-days}                                                        & 17   & \textbf{re-ut}                                                             & 30    \\ 
\hline
\textbf{dos-val}                                                           & 1775 & \textbf{freq-int-v1}                                                      & 31   & \textbf{ns}                                                                & 0     \\ 
\hline
\textbf{dos-uf}                                                            & 1679 & \textbf{freq-int-v1-ut}                                                   & 26   & \textbf{nr}                                                                & 0     \\ 
\hline
\textbf{dos-cond}                                                          & 145  & \textbf{freq-int-v2}                                                      & 20   & \textbf{qsp-val}                                                           & 30    \\ 
\hline
\textbf{max-unit-val}                                                      & 30   & \textbf{freq-int-v2-ut}                                                   & 10   & \textbf{qsp-ut}                                                            & 29    \\ 
\hline
\textbf{max-unit-uf}                                                       & 18   & \textbf{dur-val}                                                          & 1402 & \textbf{dur-ut}                                                            & 1402  \\
\hline
\end{tabular}
    }
    \caption{Slot-label distribution of PxSLU Corpus}
    \label{tab:distributionTab}
\end{table}


\bibliographystyle{model5-names}
\bibliography{lrec2022-example}


\end{document}